\documentclass{article} %
\usepackage{iclr2024,times}

\usepackage{hyperref}
\usepackage{url}

\usepackage{booktabs}
\usepackage{url}
\usepackage{times}
\usepackage{latexsym}
\usepackage{hyperref}

\usepackage{amsmath,amssymb}
\usepackage{graphicx}
\usepackage{tabularx}
\usepackage{multirow}
\usepackage{pifont}
\usepackage{soul}

\usepackage[toc,page]{appendix}
\usepackage{xcolor}
\usepackage{floatrow}
\usepackage{makecell}
\usepackage{xspace}
\usepackage{colortbl}
\usepackage{caption}
\usepackage{floatrow}

\usepackage{graphicx}
\newfloatcommand{capbtabbox}{table}[][\FBwidth]

\usepackage{arydshln}

\usepackage{minibox}
\usepackage{tabularx,ragged2e}
\usepackage{tabularx} %
\usepackage{algorithm}
\usepackage[noend]{algpseudocode}
\usepackage{bbm}

\definecolor{aquamarine}{rgb}{0.5, 1.0, 0.83}

\definecolor{summ9}{HTML}{B1DBA6}
\definecolor{summ7}{HTML}{C8EAD6}
\definecolor{summ2}{HTML}{C9E4E7}
\definecolor{leafprompt}{HTML}{E8DAEF}
\definecolor{triageprompt}{HTML}{FFE7B5}
\definecolor{HLRed}{HTML}{FFC3C3}

\newcommand{\sysname}{\textsc{MemWalker}}

\title{Walking Down the Memory Maze: Beyond Context Limit through Interactive Reading}

\author{Howard Chen\footnotemark[1] \\
Princeton University\\
\And Ramakanth Pasunuru \\ Meta AI \\
\And Jason Weston \\ Meta AI \\
\And Asli Celikyilmaz \\ Meta AI \\
}

\iclrfinalcopy %
\begin{document}

\maketitle
\renewcommand{\thefootnote}{\fnsymbol{footnote}}
\footnotetext[1]{Work done during internship at Meta AI.}

\begin{abstract}
Large language models (LLMs) have advanced in large strides due to the effectiveness of the self-attention mechanism that processes and compares all tokens at once.
However, this mechanism comes with a fundamental issue --- the predetermined context window is bound to be limited. Despite attempts to extend the context window through methods like extrapolating the positional embedding, using recurrence, or selectively retrieving essential parts of the long sequence, long-text understanding continues to be a challenge.
We propose an alternative approach which instead treats the LLM as an interactive agent,  allowing it to decide \textit{how} to read the text via iterative  prompting.
We introduce \sysname{}, a method that first processes the long context into a tree of summary nodes. Upon receiving a query, the model navigates this tree in search of relevant information, and responds once it gathers sufficient information. On long-text question answering tasks
our method outperforms  baseline approaches that use long context windows, recurrence, and retrieval. We show that, beyond effective reading, \sysname{} enhances explainability by highlighting the reasoning steps as it interactively reads the text; pinpointing the relevant text segments related to the query.
\end{abstract}

\section{Introduction}\label{sec:intro}

Large language models (LLMs) have witnessed significant advancements due to the increased model size, expanded pretraining data, and the adoption of the Transformer architecture with self-attention \citep{vaswani2017attention}. As LLMs evolve in capability, users increasingly seek to use longer input sequences during inference. This results in a growing demand in querying for information in long documents, analyzing legal or scientific papers, and managing extended conversational dialogues. 
These tasks involve consuming large amount of information, highlighting the importance of longer context processing.

Despite the rapid development, the limitation of the self-attention mechanism becomes apparent as its memory usage increases with longer sequences, consequently limiting the size of the context window. To address this, different approaches have been employed, such as designing lighter and more efficient attention schemes \citep{zaheer2020bigbird}, finetuning with extrapolated or interpolated positional embeddings \citep{press2022train, chen2023extending}, incorporating recurrence to bring forward information from preceding text segments into the next \citep{rae2019compressive,fan2020feedback,xu2022beyond}, or retrieving relevant parts of the text \citep{lewis2020retrieval,izacard2021leveraging}. However, these approaches are still limited by design. The context window, no matter how long it is extended, assumes a fixed size, and not all positions within it hold equivalent significance \citep{liu2023lost}. While recurrence can manage infinite-length sequences, it often misses out on retaining information from earlier segments.  Additionally, retrieving segments from the coherent long-text might be ineffective, given that many retrieval systems are tailored to distinguish similar but distinct documents \citep{chen2017reading}.

\begin{figure}[t]
    \centering
    \includegraphics[width=0.85\textwidth]{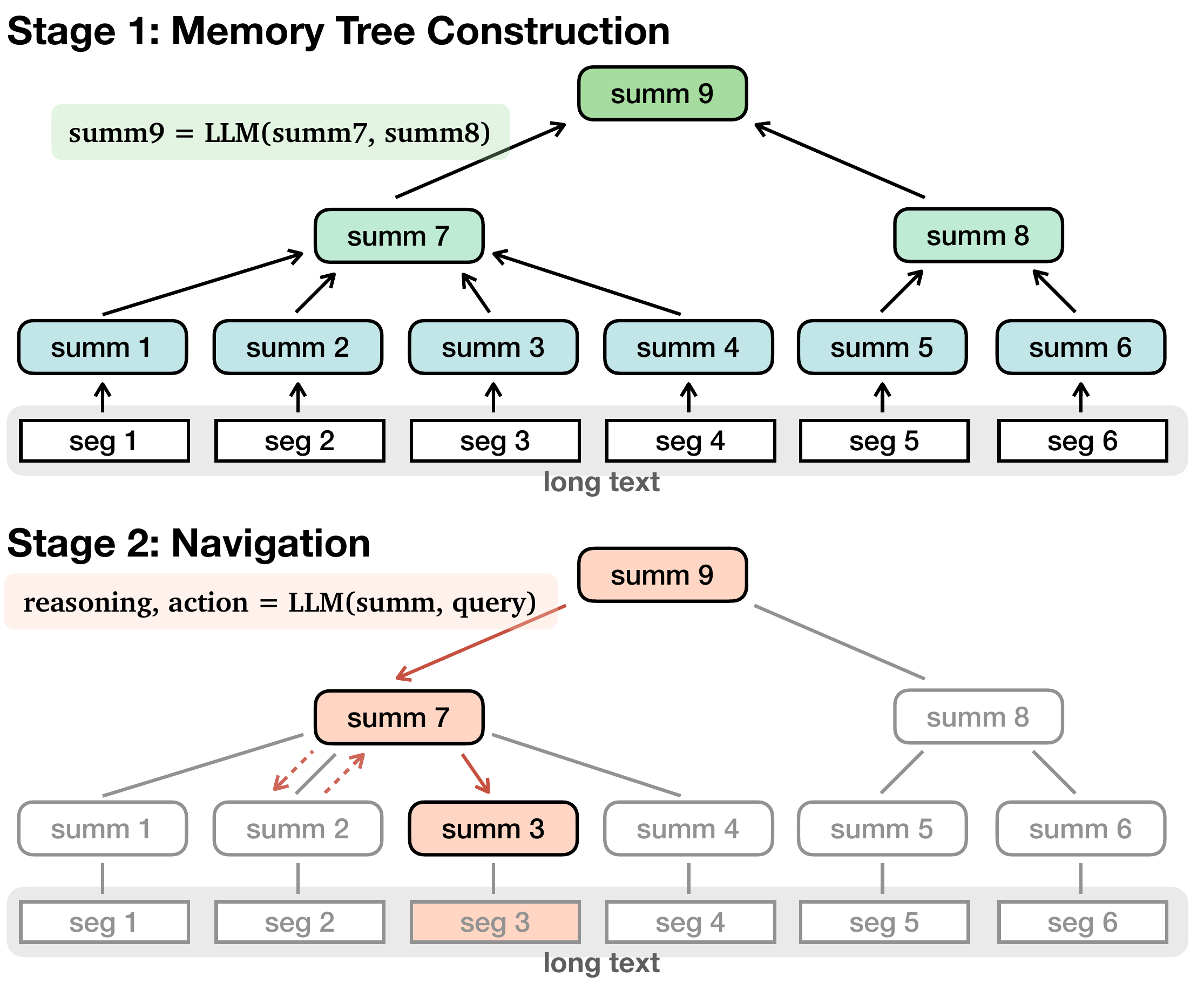}
    \caption{
        The two-stage procedure of \sysname{}. Top (stage 1): the memory tree is constructed. The long text is split into segments of a predetermined size and  each segment is first summarized into a summary node. The summary nodes are recursively summarized into higher level nodes until it reaches the root.
        Bottom (stage 2): Given a query, the LLM navigates the tree structure via iterative prompting and finds the node that contains relevant segment to form the answer. At each node, the LLM decides the action by first reasoning about the child summary nodes by sampling from the distribution $\texttt{LLM}(reasoning, action \mid summ, query)$.
        The LLM can choose the {\em revert} action to return to the parent node if it chose the wrong path or the segment at hand is irrelevant (dashed red arrow). See Table~\ref{table:examples} for a detailed example showing the LLM prompts that enable navigation.
    }
    \label{fig:teaser}
\end{figure}

To address these issues, we develop a fundamentally different approach which
treats the model with a finite context window as an interactive agent, rather than simply processing the entire sequence in one go. To this end, we introduce \sysname{}, a method that enables the model to read the long-text interactively via iterative LLM prompting. \sysname{} operates through a two-stage approach: 1) \textit{memory tree construction} and 2) \textit{navigation}. During the first stage, the long-text is segmented into small chunks that fit within the LLM's context window. The LLM then subsequently summarizes each segment into a textual summary node. These summary nodes are progressively further summarized into higher-level summary nodes, thus building a tree structure (Figure~\ref{fig:teaser}). To answer a user query, the LLM begins navigation from the tree's root node. It traverses the tree, inspecting various parts of the text to identify the path and segment relevant to answer the query. As a result, \sysname{} can go beyond the context limit, efficiently processing texts and localizing the important segments of the long-text,  without additional finetuning.

We evaluate \sysname{} on three long context question answering tasks and show superior performance against recurrence, retrieval, and %
vanilla LLM baselines. \sysname{} also outperforms other open long context systems that can take $8,000$ to $16,000$ tokens. We provide an analysis of the effectiveness of  %
\sysname{},  and show it can
reason about navigation decisions, incorporate working memory during traversal, and
recover from errors made in early navigational steps.

\section{Related Work}\label{sec:related}

\paragraph{Context window scaling.}
A straightforward approach to enable a longer context sequence is to tune the pre-trained language models and extrapolate their positional embeddings on longer text sequences \citep{press2022train, chen2023extending}.
Another direction is modified self-attention \citep{beltagy2020longformer, zaheer2020bigbird, guo2022longt5, ainslie2023colt5}.
This approach has advanced in large strides thanks to training techniques such as Flash Attention \citep{dao2022flash} that greatly reduce the memory footprint.
Despite the recent advances, this approach comes with two natural limitations: 1) to enable models on longer sequences, the model needs to be fine-tuned, incurring a non-negligible cost and 2) the attention mechanism may become less effective due to positional biases as the sequence length becomes very long \citep{liu2023lost}.

\paragraph{Recurrence.}
Recurrent architectures have been extensively studied to tackle long sequence problems, from  recurrent neural network based models \cite{hochreiter1997lstm, miller2016memorynetwork} to the modern Transformer based models \citep{dai2019transformerxl, rae2019compressive, fan2020feedback, xu2022beyond, bulatov2023rmt, chevalier2023adapting}. However, each recurrence step incurs information loss and the training objective does not guide ``how to compress'' with regard to downstream tasks. Typically this compression means that recall of older sequence information is weaker compared to recent information.

\paragraph{Retrieval.}
Retrieval systems are commonly used to select relevant documents from a large pool of documents, and have been incorporated into neural models in various ways \citep{chen2017reading,dinan2018wizard,lewis2020retrieval}. For long sequence reading, retrieval based methods typically first  embed the text segments into vector representations and retrieve them based on the query instead of feeding the entire sequence into the model such as in Fusion-in-Decoder \cite{izacard2021leveraging} or kNN variants that attend to external memory such as Memorizing Transformers \citep{wu2022memorizing}.

\paragraph{Reasoning agents.}
Instead of taking the long text as a single monolithic input, a model can act as an agent that reads part of the text and takes flexible actions. Work such as WebGPT \citep{nakano2021webgpt} and WebShop \citep{yao2022webshop} allow the model to  scroll through the internet and search for the requested answer or item. While their atomic actions allow for interactive search for relevant content, the models were not designed for understanding long and coherent texts. On the other hand, PEARL \citep{sun2023pearl} prompts the model to generate pseudo APIs for the model to call in order to focus on the right parts of the long text. However, the method operates %
within the LLM's context window, rather than being a memory-access approach that goes beyond the context limit.
Other works leveraged iterative prompting to reason and plan for long text generation tasks such as Re3 \citep{yang2022re3} and RecurrentGPT \citep{zhou2023recurrentgpt}. Self-Notes \citep{lanchantin2023selfnote} interleaved self-generating notes and the input data to perform better reasoning.
Prior to current LLMs, LSTMs were also applied to searching through document structures (titles, subsections) \cite{geva2018learning}. 
Recursive tree structure has also been explored  in the context of summarization of long text such as books in  \citep{wu2021recursively}, but was not used for memory navigation in that work.

\section{\sysname{}: An interactive reader}\label{sec:method}\label{interactivereading}

\begin{table*}[t]
    \centering
    \caption{
        Example trajectory from the QuALITY dataset. The LLM first sees the content of the children nodes at the root node (\texttt{\sethlcolor{summ9}\hl{summ 9}} in Figure~\ref{fig:teaser}) and generates the response (takes action 0 to enter \texttt{\sethlcolor{summ7}\hl{summ 7}}). When arriving at the leaf node (\texttt{\sethlcolor{summ2}\hl{summ 2}}), the LLM determines that there is not enough information, therefore takes the action to revert (action -1) to the parent node.
        After hopping back-and-forth between nodes, the LLM commits to a Leaf node (\texttt{\sethlcolor{summ2}\hl{summ 3}}) and answers the question. Yellow indicates \sethlcolor{triageprompt}\hl{triage prompt} and purple indicates \sethlcolor{leafprompt}\hl{leaf prompt} described in \S\ref{sec:method}. Text after // denotes comments that are not processed by the LLM.
    }
    \label{table:examples}
    \resizebox{1.0\columnwidth}{!}{
        \begin{tabular}{lp{15cm}}
        \toprule
        \multicolumn{2}{c}{\textbf{Navigation Trajectory}} \\
        \midrule
        \texttt{\sethlcolor{summ9}\hl{summ 9}} & \sethlcolor{triageprompt}\hl{The following passage(s) are the summaries of the different parts of a story.} \newline
        \sethlcolor{triageprompt}\hl{To answer the question:} \textit{Why did Ro change his mind about the people on Mars being backwards?} \newline
        \sethlcolor{triageprompt}\hl{Which of the following summary is MOST LIKELY to contain information about the answer?} \newline
        \sethlcolor{triageprompt}\hl{First provide reasoning to compare the summaries before you make the decision.} \newline \newline
        Summary 0: The story is set on Mars and follows the adventures of Ro, [...] // \texttt{\sethlcolor{summ7}\hl{summ 7}} \newline
        Summary 1: Ro, a young Martian, is climbing down a cliff to rescue [...] // \texttt{\sethlcolor{summ7}\hl{summ 8}} \newline
        \newline
        \sethlcolor{triageprompt}\hl{Relpy with the passage number as your action.} \newline
        \sethlcolor{triageprompt}\hl{You MUST choose one summary number and you should reply with the following format:} \newline
        \sethlcolor{triageprompt}\hl{\#\#\#\#\#\#\#\#\#\#\#\#\#\#\#\#\#\#\#\#\#\#\#\#\#\#\#\#\#\#\#\#\#\#\#} \newline
        \sethlcolor{triageprompt}\hl{Reasoning: ...} \newline
        \sethlcolor{triageprompt}\hl{Action: 0 / 1 / 2, ...} \newline
        \sethlcolor{triageprompt}\hl{\#\#\#\#\#\#\#\#\#\#\#\#\#\#\#\#\#\#\#\#\#\#\#\#\#\#\#\#\#\#\#\#\#\#\#} \\
        Response & Reasoning: Summary 0 is most likely to contain information about why Ro changed his mind about the people on Mars being backwards, as it mentions Ro's interactions with the people from Earth and their advanced method of communication. \newline Action: 0 \\
        \midrule
        \multicolumn{2}{l}{LLM enters \texttt{\sethlcolor{summ7}\hl{summ 7}} then enters \texttt{\sethlcolor{summ2}\hl{summ 2}}} \\
        \midrule
        \texttt{\sethlcolor{summ2}\hl{summ 2}} & \sethlcolor{leafprompt}\hl{Read the text in triple quotes and answer a question:} \newline
        \sethlcolor{leafprompt}\hl{Story background information:} Ro, a young Martian who has returned from his travels to find his home overrun by the Oan, a tribe of rat-like creatures. [...] // $\texttt{[WORKING\_MEMORY]}$ carried from \texttt{\sethlcolor{summ7}\hl{summ 7}} \newline
        \sethlcolor{leafprompt}\hl{Main text:} In his left hand and under his armpit Ro carried stones. They were of a good weight and would make short work of any Oan who was foolish enough to cross his path. [...] // \texttt{seg 2} \newline \newline
        Why did Ro change his mind about the people on Mars being backwards? \newline
        (A) He realized that despite human's technological advancements, they have over-complicated marriage. \newline
        (B) He realized that while the humans are physically vulnerable without their weapons, the red people have formidable strength in their arms. \newline
        (C) He realized that human males suppress public affection when they are intimidated by other males, whereas male Martians don't hide their affection. \newline
        (D) He realized that male humans were petty and even brute when it came to rivalry over women, whereas male Martians were much more civilized. \newline
        \sethlcolor{leafprompt}\hl{If the answer CANNOT be inferred from the text above, reply with action -1.} \newline
        \sethlcolor{leafprompt}\hl{If the answer CAN be inferred from the text above, reply with action -2, and also provide your reasoning, and the final answer.} \newline
        \sethlcolor{leafprompt}\hl{You are ONLY allowed to reply with action -2 or -1.} \newline
        \sethlcolor{leafprompt}\hl{Your should reply with the following format:} \newline
        \sethlcolor{leafprompt}\hl{\#\#\#\#\#\#\#\#\#\#\#\#\#\#\#\#\#\#\#\#\#\#\#\#\#\#\#\#\#\#\#\#\#\#\#} \newline
        \sethlcolor{leafprompt}\hl{Reasoning: ...} \newline
        \sethlcolor{leafprompt}\hl{Action: -2 or -1} \newline
        \sethlcolor{leafprompt}\hl{Answer: (A) ...} \newline
        \sethlcolor{leafprompt}\hl{\#\#\#\#\#\#\#\#\#\#\#\#\#\#\#\#\#\#\#\#\#\#\#\#\#\#\#\#\#\#\#\#\#\#\#} \\
        Response & Reasoning: The text does not explicitly mention Ro changing his mind about the people on Mars being backwards. Therefore, the answer cannot be inferred from the text. \newline Action: -1 \\
        \midrule
        \multicolumn{2}{l}{LLM reverts to \texttt{\sethlcolor{summ7}\hl{summ 7}} then enters \texttt{\sethlcolor{summ2}\hl{summ 3}}} \\
        \midrule
        \texttt{\sethlcolor{summ2}\hl{summ 3}} & \sethlcolor{leafprompt}\hl{[...] Main text:} ``Last night I thought that we on Mars are backward. Now I'm not so sure.'' {[...]}  // \texttt{seg 3} \sethlcolor{leafprompt}\\
        Response & Reasoning: Ro initially sees Earth's customs as backward compared to Mars, However, after discussing [...] \newline
        Action: -2 \newline
        Answer: (A) \\
        \bottomrule
        \end{tabular}
    }
\end{table*}

We study tasks related to long-context question answering -- given a long-text $x$ and a query $q$, the  model aims to generate the response $r$. %

\sysname{} follows two steps: 1) \textit{memory tree construction}, where the long-context is broken down into a tree data structure. This construction does not depend on the query, and can hence be computed in advance if the sequence data is available beforehand.  2) \textit{navigation}, in which the model navigates this structure upon receiving a query, gathering information to craft a suitable response. \sysname{}~assumes access to an underlying LLM, and both construction and navigation are achieved through iterative LLM prompting.

\paragraph{Memory tree construction.}
\sysname{} first creates a tree data structure, $\mathcal{T}(x)$, from the long-text $x$. Each node is represented by text that encapsulates the summaries of all its child nodes below it. Specifically, the long-text $x$ is divided into segments $(c_1, \dots, c_n)$. The LLM then summarizes each segment into a summary at the first level, represented as $s^{l=1}_i = \texttt{LLM}(c_{\leq i})$, $i = 1 .. n$. The initial summary nodes are subsequently summarized further into higher level nodes, $s_j^{l+1} = \texttt{LLM}(s^l_i, \dots, s^l_{i + M_t})$ where $M_t$ denotes the number of nodes in the $t$-th grouping at level $l$. This process continues until the topmost root node, $s^L$ is generated. The complete tree generation process is illustrated in Figure~\ref{fig:teaser}. Summarization is performed using LLM prompting.  We include the prompts for memory tree construction in Appendix~\ref{app:prompts}. %

\paragraph{Navigation.}
The language model, upon receiving a query $q$, starts from the root node $s^L$ to navigate the tree to generate the response $r$. 
At node $s^{l}$ that the LLM traverses, it observes the summaries of the nodes one level below $\{ s^{l-1}_i, \dots, s^{l-1}_{i+M_t} \}$. The LLM decides among $\lvert M_t \rvert + 1$ actions --- choosing one of the child nodes to further inspect, or to revert to the parent node. At leaf node $s^{l=1}_i$, the LLM can decide one of two actions: \textit{commit} to the leaf node and respond to the query or \textit{revert} to the parent node ($s^{l+1}_j$) if the information in the leaf node (i.e., $c_i$) is insufficient.
To make a navigation decision, we can also ask the LLM (via prompting) to 
 first generate a reason in natural language to justify the action, followed by the action choice itself. Specifically, at each node, the model generates a response $r \sim \texttt{LLM}(r \mid s, q)$ where the response is either of the two tuples: 
1) $r = (reasoning, action, answer)$ when the LLM is at a leaf node or 2) $r = (reasoning, action)$ when the LLM is at non-leaf nodes.

\paragraph{Navigational prompt design.}

We enable LLM navigation through zero-shot 
prompting. 
Our method requires two types of prompt:
1) \textit{\sethlcolor{triageprompt}\hl{triage prompt}} and 2) \textit{\sethlcolor{leafprompt}\hl{leaf prompt}} (highlighted in Table~\ref{table:examples}).
Triage prompt contains the the query, the summaries of the children nodes, and instructions for the LLM to follow. Triage prompt is used at non-leaf nodes.
Leaf prompt contains the content of the segment, the query (and options), and instructions that ask the LLM to either generate the answer or revert to the parent node. Both the triage prompt and leaf prompt specify an output format that the LLM needs to follow. Failure to conform to the format results in invalid actions and the LLM is required to regenerate. If the LLM fails to generate parsable output three consecutive times, the navigation terminates and returns ``no answer''.

\paragraph{Working memory.} As the LLM traverses the tree, it can keep information throughout the navigation trajectory and add it to the context. Formally, the LLM generates the response $r \sim \texttt{LLM}(r \mid s, q, m)$ where the extra working memory $m \in \{\O\} \cup \{ (s_i, s_{i+1}, \dots) \}$ is either empty or consists of contents from previously visited nodes. We truncate the working memory such that they can fit in the LLM's context window.\footnote{Further summarizing the working memory as it accumulates would be an alternative approach, which we have not explored in this study.} Table~\ref{table:examples} illustrates the way working memory is added via  \texttt{[WORKING\_MEMORY]} in the prompt.

\section{Experimental Setup}\label{sec:exp}

\subsection{Datasets \& Evaluation}
We use three datasets: QuALITY, SummScreenFD, and GovReport from the SCROLLS benchmark \citep{shaham2022scrolls}.
We report accuracy for all datasets.

\paragraph{QuALITY.} QuALITY is a multiple choice question answering dataset collected by \cite{pang2022quality}. The dataset contains long-form stories sourced from Project Gutenberg and questions annotated by human annotators. We use a subset of $187$ examples for our experiments.

\paragraph{SummScreenFD.} SummScreenFD \citep{chen2022summscreen} is a dataset of TV and movie scripts in the form of dialogues among actors originally designed for summarization. We repurpose the dataset into a question answering task where the original provided ground truth summary text is used to generate a ``who'' question using Stable Beluga 2, with answers then checked by a human expert. The question paired with the original long text becomes %
the repurposed QA task of $306$ examples.

\paragraph{GovReport.} The GovReport dataset aggregates documents from Congressional Research Service and the U.S. Government Accountability Office together with summaries provided by experts \citep{huang2021efficient}. We repurpose the dataset into a question answering dataset of $101$ examples the same way as for SummScreenFD.

All three datasets feature long contexts per example of varying length -- some shorter examples, and some longer sequences. We therefore both report results on the original dataset, and also report on a subset of each task containing only longer sequences, to better evaluate memory access in the harder, longer context case.  The thresholds are above 8, 000 tokens for QuALITY, 6, 000 tokens for SummScreenFD, and 12, 000 tokens for GovReport.

\subsection{Model}
We use Stable Beluga 2 \citep{StableBelugaModels} as the base LLM for the majority of our experiments, as it provides  state-of-the-art performance compared to several other LLM variants, as we will show. 
Stable Beluga 2 is an instruction-tuned model
built on top of  70B LLaMA-2\citep{touvron2023llama2}, where the finetuning does not overlap with our evaluation tasks.
It has a maximum $4,096$ token context length.
We use the model in a zero-shot prompting fashion without further fine-tuning or in-context few shot examples for our tasks. We use top-$p$ sampling for both memory tree construction as well as generating action and reasoning for navigation. We set the maximum number of nodes $\max_t M_t = 8, 5, 8$ and segment size $\lvert c \rvert = 1000, 1000, 1200$ for QuALITY, SummScreenFD, and GovReport respectively.

\subsection{Baselines}
We compare with three baselines memory techniques all based on the same underlying LLM, Stable Beluga 2: 1) full context window, 2) recurrence, and 3) retrieval.
The full context window baselines utilize the full $4,096$ tokens to process both the long input text and generation. Since the instances in the dataset often exceed the context limit, we perform truncation of the length to the right (most recent) or left (least recent) of the text as the input, as evaluate both approaches. For retrieval, we use Contriever \citep{izacard2022unsupervised} to select segments from the long context based on the query. The highest scored segments are concatenated as the input context to the LLM until they fill the context. 
Finally, we implement a baseline that recurrently carries information from previous segment tokens to the current one through summarization \citep{xu2022beyond}, where each segment is $2,500$ tokens and the maximum summary size is $500$ tokens.

\section{Results \& Analysis}\label{sec:results}

\begin{table*}[t]
    \centering 
    \caption{
        Results on the three question answering tasks, reporting test accuracy. Orig. denotes using the entire dataset and Long denotes the subset of longer sequences. 
        Top: comparison to open long context models. Bottom: baselines and \sysname{} performance, with all methods using the underlying Stable Beluga 2 LLM with a maximum $4,096$-token context length. \sysname{}~outperforms all other systems on longer sequences.
    }
    \resizebox{0.8\columnwidth}{!}{
        \begin{tabular}{lccc}
        \toprule
         & QuALITY & SummScreenFD & GovReport \\
         & Orig. / Long & Orig. / Long & Orig. / Long \\
        \midrule
        MPT 13B (8k) & 44.4 / 47.3 & 65.0 / 63.5 & 44.6 / 43.8\\
        LongChat 13B (16k) & 43.3 / 48.4 & 62.4 / 61.1 & 54.5 / 52.1 \\
        \midrule
        Recurrence & 51.3 / 56.0 & 47.7 / 45.4 & 35.6 / 33.8 \\
        Retrieval & 63.1 / 64.8 & 63.7 / 62.2 & 54.0 / 52.1 \\
        Full Context (keep left) & 56.7 / 64.8 & 62.7 / 62.7 & \textbf{59.4} / 56.3 \\
        Full Context (keep right) & \textbf{70.1} / 72.5 & 64.7 / 63.1 & 50.5 / 50.0 \\
        \sysname{} & 67.4 / \textbf{73.6} & \textbf{67.3} / \textbf{64.5} & \textbf{59.4} / \textbf{60.4} \\
        \bottomrule
        \end{tabular}
    }
    \label{table:main_results}
\end{table*}

\paragraph{Main results.}
Table~\ref{table:main_results} shows comparisons between \sysname{} and other baselines. \sysname{} outperforms both the recurrence  baseline across all tasks by a large margin. This shows the limitation of recurrence, where  relevant information to the query is lost after several steps. 
 \sysname{}  also outperforms retrieval where the segments are from a coherent long story instead of separate documents.
On these tasks, the full context baselines can perform well in the ``Original'' task setting, which can contain relatively shorter sequences, although choosing either left or right truncate for best performance seems to be dataset dependent.  Still, \sysname{} achieves higher performance in the Original setting against the Full Context baselines except for the keep right variant on QuALITY and the keep left variant on GovReport, likely due to the positional bias in the dataset where relevant segment often appears at the beginning or the end of the text. However, on the Long version of all three tasks \sysname{} outperforms all baselines, that is it shows strong performance when memory access becomes more critical.
\sysname{} also outperforms other publicly available models, including LongChat \citep{longchat2023} and MPT \citep{mosaic2023introducing}.

\begin{figure}[t!]
    \vspace{10pt}
    \centering
    \includegraphics[width=1.0\textwidth]{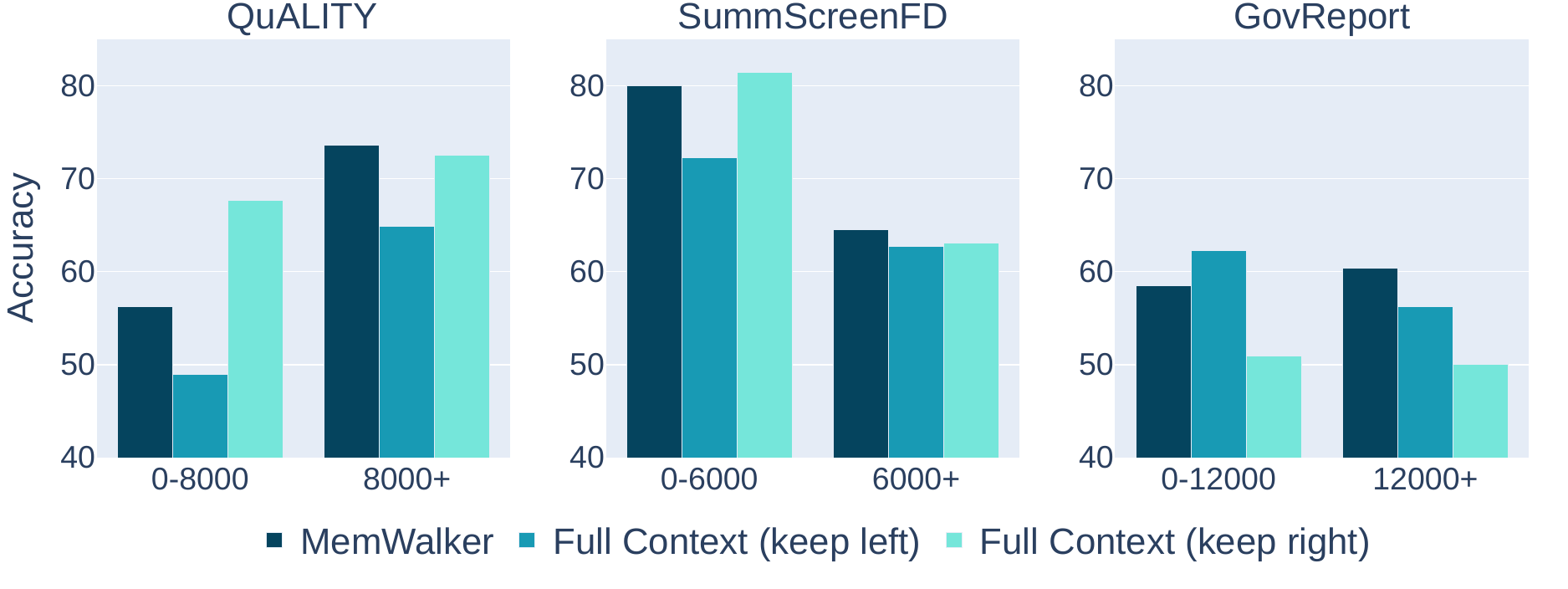}
    \caption{
        Performance breakdown by context length (in tokens). Each dataset is thresholded into two bucket of equal sizes. \sysname{} outperforms full context baselines (truncated either left or right, when the sequence does not fit) on longer context sequences, for all three tasks.
    }
    \label{fig:perf_breakdown_by_length} 
    \vspace{-9pt}
\end{figure}

\paragraph{\sysname{} improves performance on long sequences.}
We provide a breakdown of performance by input sequence length for each task in Figure~\ref{fig:perf_breakdown_by_length}.
\sysname{} is not advantageous over Full Context (with truncation left or right) baselines when the text length is short, but outperforms both types of truncation for all tasks for longer sequences. The benefit of interactive reading emerges after the text length is suitably large, i.e. showing better performance once the sequence length is sufficiently larger than the LLM context length of $4,096$.

\begin{table*}[t]
    \centering
    \vspace{-5pt}
    \resizebox{1.0\columnwidth}{!}{
        \begin{tabular}{lccc}
        \toprule
         & QuALITY & SummScreenFD & GovReport \\
         & Acc. / Valid Action (\%) & Acc. / Valid Action (\%) & Acc. / Valid Action (\%) \\
        \midrule
        LLaMA 2 Chat (13B)    & 39.6 / ~~73.2 & 20.9 / ~~75.5 & 15.8 / ~~69.0 \\
        ~~~~ w/o reasoning    & 48.1 / ~~97.4 & 25.8 / ~~95.8 & 21.8 / ~~93.1 \\
        \midrule
        LLaMA 2 Chat (70B)    & 52.0 / ~~86.1 & 55.6 / ~~99.5 & 41.6 / ~~97.8  \\
        ~~~~ w/o reasoning    & 59.9 / 100.0 & 58.5 / 100.0 & 42.6 / 100.0 \\
        \midrule
        Stable Beluga 2 (70B) & 67.4 / ~~92.5 & 67.3 / ~~95.1 & 59.4 / ~~97.0\\
        ~~~~ w/o reasoning    & 66.8 / 100.0 & 64.1 / ~~90.5 & 52.5 / ~~98.2 \\
        \bottomrule
        \end{tabular}
    }
    \caption{
        \sysname{} performance using different underlying LLMs with different reasoning capabilities, and an ablation on their reason justification component when making a navigation decision (``w/o reasoning'' simply predicts the action, with no reason generated, see e.g. \autoref{table:examples}). Valid Action shows the percent of generated actions that are a valid navigation action. We find that the strongest performing LLM (Stable Beluga 2) benefits from reasoning with improved accuracy, while weaker performing LLMs do not (get worse in terms of accuracy and valid actions).  
    }
    \label{table:model_reason_capability}
\end{table*}

\paragraph{Reasoning capability is essential for memory tree navigation.}
The effectiveness of \sysname{} is highly dependent on the underlying LLM's reasoning capability. 
For each navigation decision, we employ an LLM prompt that requires the LLM to first generate a reason in natural language that justifies the following predicted action, see \autoref{table:examples}.
We show in Table~\ref{table:model_reason_capability} how reasoning impacts performance by comparing Llama 2 Chat (13B and 70B parameter variants) and Stable Beluga 2 (70B)  with and without the reasoning justification by removing the 
line ``First provide reasoning \dots before you make your decision'' from the prompt. With the smaller, less capable models (13B), the performance lags behind 70B models by a large margin due to its inability to follow instructions. %
In fact, asking for reasoning justifications for weaker models 
{\em decreases performance}, presumably due to their inability to generate and make use of such reasons.
Stable Beluga 2 outperforms Llama 2 Chat for the same LLM size, and also displays heightened reasoning ability.
For Stable Beluga 2, asking for reasoning justification {\em improves performance} across all tasks.
This highlights the main characteristic of \sysname{}: if an LLM passes a critical reasoning ability threshold, it can reason about a long input in multiple rounds without errors cascading quickly across rounds. For weaker LLMs that cannot make good navigation decisions, 
errors could compound and overall performance suffers.
As LLMs will only improve in reasoning ability over the coming years, we expect methods like \sysname~will become more and more effective.

\begin{figure}[t!]
    \centering
    \includegraphics[width=.8\textwidth]{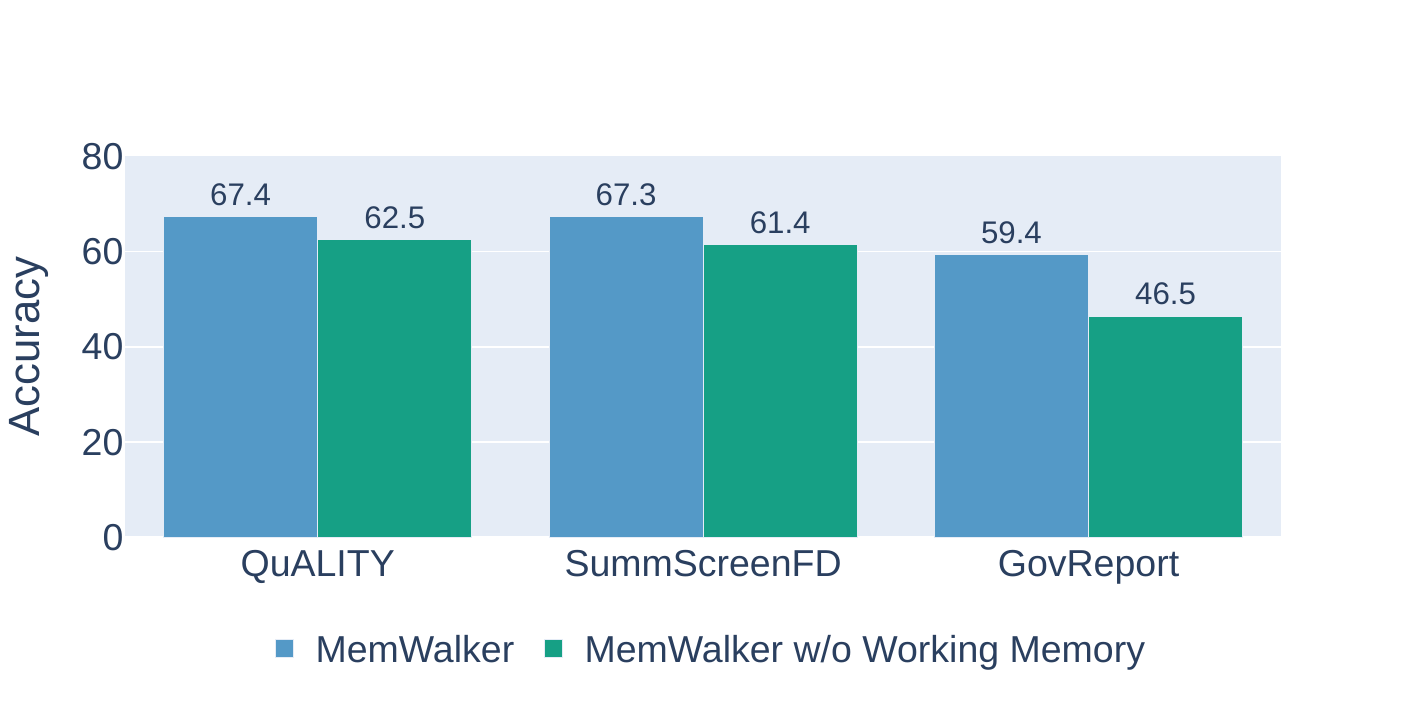}
    \caption{
        \sysname{} performance comparisons between using working memory and without (i.e., the LM only looks at the content of the children memory tree nodes, rather than memory from all the nodes it has traversed). Inclusion of working memory yields large gains.
    }
    \label{fig:working_memory} 
    \vspace{-8pt}
\end{figure}
\begin{table}[t]
\RawFloats
\begin{minipage}{.47\linewidth}
    \centering
    \caption{\sysname{}~navigation analysis. Stray ratio: percentage of paths that contain the \textit{revert} action. Recovery Rate: percentage of stray paths that recover and answer the query correctly.}
    \label{table:recovery}
    \begin{tabular}{lcc}
        \toprule
         & Stray & Recovery \\
         & Ratio & Rate \\
        \midrule
        QuALITY & 15.0 & 70.0 \\
        SummScreenFD & 18.6 & 59.6 \\
        GovReport & 18.8 & 79.0 \\
        \bottomrule
    \end{tabular}
\end{minipage}
\begin{minipage}{.52\linewidth}
    \centering
    \includegraphics[width=0.9\textwidth]{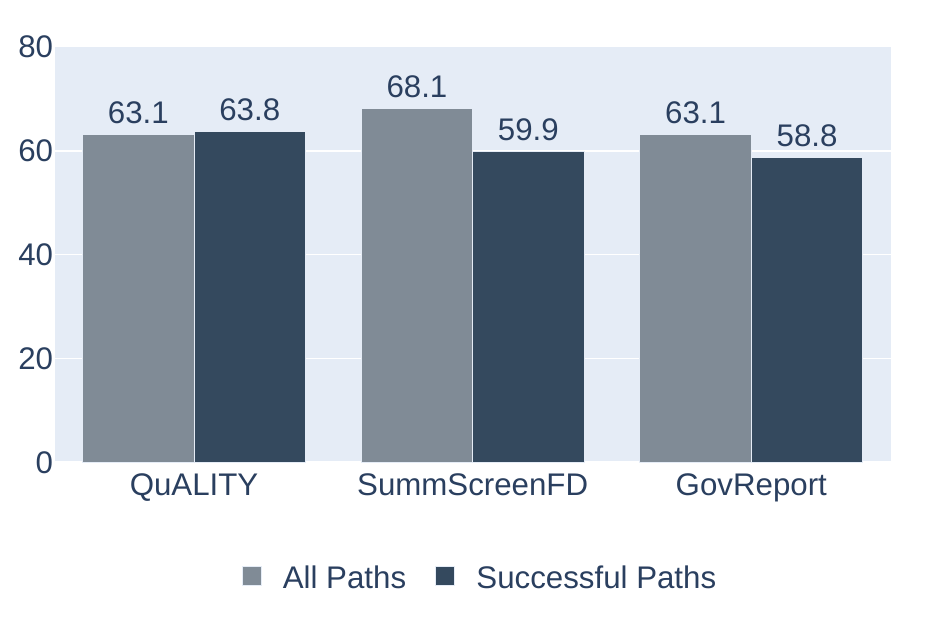}
    \captionof{figure}{Percentage comparison of total tokens processed against the tokens of the original example between all paths vs. successful paths.}
    \label{fig:efficient_reading}
\end{minipage}
\end{table}
\vspace{-7pt}

\paragraph{Navigating the memory tree requires working memory.}
As \sysname{} makes decisions to traverse the memory tree and read  relevant segments, it might lose sight of the overall context. The model thus carries information from the nodes along the navigation path as working memory, where the content of the working memory updates as the model selects the next path. 
We evaluate the performance of \sysname{}~with and without working memory, with results given in Figure~\ref{fig:working_memory}.
We find a significant performance
degradation without working memory across all tasks, with a 5--13\% drop in accuracy,
showing the importance of this component.

\paragraph{\sysname{} can recover from stray paths.}
As \sysname{}~navigates the memory tree, it needs to not only find the path towards the most pertinent segments, but also potentially to recover from traversal errors should they occur. We report recovery statistics in  Table~\ref{table:recovery}.  
\sysname{} executes a revert navigation action (and hence changes path) for around $15\%-20\%$ of examples, but of those examples can recover and get those examples correct $70\%$ of the time for QuALITY, $\sim60\%$ for SummScreenFD, and $\sim80\%$ for GovReport.

\paragraph{\sysname{} enables efficient reading.}
Since \sysname{} determines which parts of the long text it needs to read, the effective content that needs to be read may be  smaller than the entire sequence. We report the percentage of the long context read  averaged over all examples, for each of the three tasks, in Figure~\ref{fig:efficient_reading}. We find that between only 63\%-69\%  of the text on average needs to be read to answer the question including the content of the tree nodes. Among successful paths, the reading required further reduces to 59\% - 64\%.

\begin{figure}[t!]
    \centering
    \includegraphics[width=0.8\textwidth]{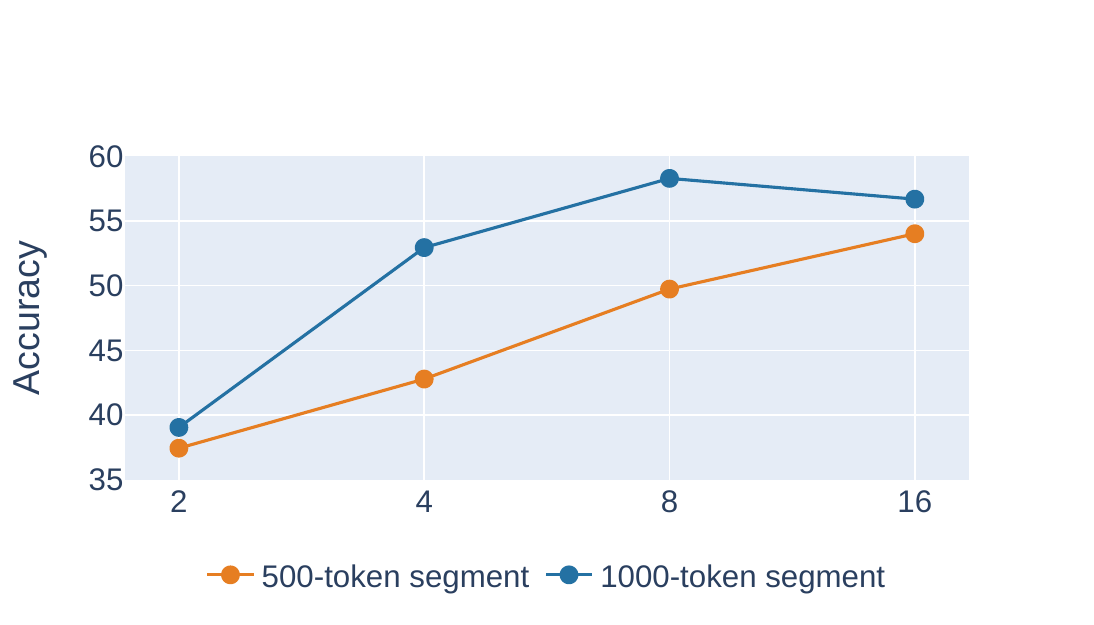}
    \caption{
        Performance trade-off of different memory construction configurations on QuALITY. x-axis: maximum number of nodes that can be connected to a parent node. Red: summarizing $1,000$-token segments. Blue: summarizing $500$-token segments.
    }
    \label{fig:tradeoff}
    \vspace{-12pt}
\end{figure}

\paragraph{Memory tree construction trade-off.}
A fundamental trade-off arises as we construct the memory tree --- summarizing larger segments compresses more information into a node to reduce the depth of the tree, but risks losing fidelity of the content. Similarly, connecting many lower level nodes to the upper one can help flatten the tree, yet render the navigation task harder for the LLM at each node.
Figure~\ref{fig:tradeoff} shows  the performance of different configurations of the memory tree on QuALITY. Summarizing larger segments is generally more beneficial than smaller segments as well as connecting more children nodes to the parent. However, the performance plateaus as the maximum number of nodes increases, showing the trade-off with respect to how much information can be packed into the nodes during memory tree construction.

\section{Conclusion}\label{sec:conclusion}
\vspace{-5pt}
We propose \sysname{}, an interactive reading agent 
which uses iterative LLM prompting to decide which part of the content should be read closely based on its own reasoning.
Our approach first builds a structured memory given long context sequence data, and then makes navigation decisions of the pertinent parts to read given a query.  
Our method shows superior performance against 
a number of baselines including various long context length models, retrieval and recurrence baselines,
in particular for longer sequence tasks.
Detailed analysis highlights a number of important factors, including our method's ability to reason about navigation decisions, ability to revert navigation to a different path when necessary, and incorporation of a working memory.
Future work should explore many new directions that \sysname{}~opens up, in particular its application to different data structures other than trees, and finetuning its performance specific to the interactive reading goal.

\section{Limitations}\label{sec:limitation}
\vspace{-5pt}
\sysname{} exhibits three major limitations. First, the memory tree generation might not scale too well if the sequence's length becomes extremely long. The increase in sequence length entails more nodes in the tree and hence renders the tree construction process onerous. Workaround such as trading off the granularity of the summary in exchange for speed might be viable. Nonetheless, the issue of scaling remains a limit. In this setting it may make sense to generalize \sysname{} to a combination of tree and hash \cite{bawa2005lsh} or other alternative data structure, whilst retaining its travesersal ability via LLM prompting. 
Second, \sysname{} only works when the LLM exhibits a strong enough reasoning capability, which according to our experiments is required to be large (over 70B) and instruction-tuned. If the reasoning capability falls short, the error compounds and the method would fail. Enabling a smaller model that can perform a similar instruction following procedure could be useful for scaling the method. This could be made possible by removing the following third limitation.
Third, \sysname{} only uses zero-shot prompting and does not leverage fine-tuning to further improve the interactive reading capability. This could be done, for example, by performing interactive reading and collect the successful paths for further fine-tuning.

\bibliography{ref}
\bibliographystyle{iclr2024}
\newpage
\clearpage
\appendix
\section{Appendix}

\subsection{Prompts}\label{app:prompts} 

We provide full prompts for both memory tree construction and navigation described in \S\ref{interactivereading}.

\subsubsection{Memory Tree Construction Prompts}
We use two prompts for memory tree construction (the construction component of Table~\ref{table:prompts}). The first (leaf) instructs the LLM to summarize the text segment into a comprehensive summary. After this step,  the segments are grouped and summarized into non-leaf node summaries. The summaries (\texttt{[CHILD\_SUMM\_NODE\_0]}, \texttt{[CHILD\_SUMM\_NODE\_1]}, \dots, \texttt{[CHILD\_SUMM\_NODE\_N]}) are grouped and concatenated as the summary content of their parent node. During this process, if the concatenated summaries exceed the predetermined length, the second construction prompt is used to further summarize the text (i.e., \texttt{[SUMMARIES]}) for the parent node.

\subsubsection{Navigation Prompts}
We use two navigation prompts (triage and leaf) as described in \S\ref{sec:method}. We show the general prompt template in the navigation stage of Table~\ref{table:prompts}.

\subsection{Examples}\label{app:examples}
We provide an extra navigation example in Table~\ref{table:extra_examples}.

\clearpage
\begin{table*}[t]
    \centering
    \resizebox{1.0\columnwidth}{!}{
        \begin{tabular}{lp{15cm}}
        \toprule
        Stage & Prompt \\
        \midrule
        Construction (leaf) & \texttt{[TEXT\_OF\_SEGMNET]}. Summarize the above text comprehensively into a fluent passage. \\
        \midrule
        Construction (non-leaf) & \texttt{[SUMMARIES]}. Compress each summary into a much shorter summary. \\
        \midrule
        Navigation (triage) & \sethlcolor{triageprompt}\hl{The following passage(s) are the summaries of the different parts of a story.} \newline
        \sethlcolor{triageprompt}\hl{To answer the question:} \texttt{[QUERY]} \newline
        \sethlcolor{triageprompt}\hl{Which of the following summary is MOST LIKELY to contain information about the answer?} \newline
        \sethlcolor{triageprompt}\hl{First provide reasoning to compare the summaries before you make the decision.} \newline
        Summary 0: \texttt{[CHILD\_SUMM\_NODE\_0]} \newline
        Summary 1: \texttt{[CHILD\_SUMM\_NODE\_1]} \newline
        \dots \newline
        Summary N: \texttt{[CHILD\_SUMM\_NODE\_N]} \newline
        
        \sethlcolor{triageprompt}\hl{Relpy with the passage number as your action.} \newline
        \sethlcolor{triageprompt}\hl{You MUST choose one summary number and you should reply with the following format:} \newline
        \sethlcolor{triageprompt}\hl{\#\#\#\#\#\#\#\#\#\#\#\#\#\#\#\#\#\#\#\#\#\#\#\#\#\#\#\#\#\#\#\#\#\#\#} \newline
        \sethlcolor{triageprompt}\hl{Reasoning: ...} \newline
        \sethlcolor{triageprompt}\hl{Action: 0 / 1 / 2, ...} \newline
        \sethlcolor{triageprompt}\hl{\#\#\#\#\#\#\#\#\#\#\#\#\#\#\#\#\#\#\#\#\#\#\#\#\#\#\#\#\#\#\#\#\#\#\#} \\
        \midrule
        Navigation (leaf) & \sethlcolor{leafprompt}\hl{Read the text in triple quotes and answer a question:} \newline
        \sethlcolor{leafprompt}\hl{Story background information:} \texttt{[WORKING\_MEMORY]} \newline
        \sethlcolor{leafprompt}\hl{Main text:} \texttt{[TEXT\_OF\_SEGMNET]} \newline \newline
        \texttt{[QUERY]} \newline
        \texttt{[OPTIONS]} \newline \newline
        \sethlcolor{leafprompt}\hl{If the answer CAN be inferred from the text above, reply with action -2, and also provide your reasoning, and the final answer.} \newline
        \sethlcolor{leafprompt}\hl{You are ONLY allowed to reply with action -2 or -1.} \newline
        \sethlcolor{leafprompt}\hl{Your should reply with the following format:} \newline
        \sethlcolor{leafprompt}\hl{\#\#\#\#\#\#\#\#\#\#\#\#\#\#\#\#\#\#\#\#\#\#\#\#\#\#\#\#\#\#\#\#\#\#\#} \newline
        \sethlcolor{leafprompt}\hl{Reasoning: ...} \newline
        \sethlcolor{leafprompt}\hl{Action: -2 or -1} \newline
        \sethlcolor{leafprompt}\hl{Answer: ...} \newline
        \sethlcolor{leafprompt}\hl{\#\#\#\#\#\#\#\#\#\#\#\#\#\#\#\#\#\#\#\#\#\#\#\#\#\#\#\#\#\#\#\#\#\#\#} \\        
        \bottomrule
        \end{tabular}
    }
    \caption{
        Prompts used for the memory tree construction stage and the navigation stage. For the memory construction stage, \texttt{[TEXT\_OF\_SEGMENT]} is filled with the segment text at the leaf nodes. \texttt{[SUMMARIES]} is the concatenated summaries from the child nodes and will be further summarized if it exceeds the predetermined length. For navigation, \texttt{[QUERY]} is the query, \texttt{[OPTIONS]} are the multi-choice options (only in QuALITY), \texttt{[CHILD\_SUMM\_NODE\_n]} represents the summary text of the $n$-th child node, and \texttt{[WORKING\_MEMORY]} is the information carried from previous nodes.
        Yellow indicates \sethlcolor{triageprompt}\hl{triage prompt} and purple indicates \sethlcolor{leafprompt}\hl{leaf prompt}, as described in \S\ref{sec:method}.
    }
    \label{table:prompts}
\end{table*}
\begin{table*}[t]
    \centering
    \resizebox{1.0\columnwidth}{!}{
        \begin{tabular}{lp{15cm}}
        \toprule
        \multicolumn{2}{c}{\textbf{Navigation Trajectory}} \\
        \midrule
        Node 1 & The following passage(s) are the summaries of the different parts of a story. \newline
        To answer the question: \textit{Who invited Michael to his business school as a guest speaker?} \newline
        Which of the following summary is MOST LIKELY to contain information about the answer? \newline
        First provide reasoning to compare the summaries before you make the decision. \newline
        Summary 0: In the text, Michael and Ryan are on their way to give a speech at a business school. [...]  \newline
        Summary 1: Michael is reminiscing about his college days and suggests playing Frisbee with a college student. [...] \newline
        Summary 2: Michael Scott is giving a presentation to a group of business students, attempting to explain  [...] 
        Summary 3: In the text, there is a scene where a bat is found in the office and employees react differently to its presence.  [...] 
        \newline
        Summary 4: Pam, an artist, has an art show featuring her paintings. Roy compliments her art [...] \newline
        Relpy with the passage number as your action. \newline
        You MUST choose one summary number and you should reply with the following format: \newline
        \#\#\#\#\#\#\#\#\#\#\#\#\#\#\#\#\#\#\#\#\#\#\#\#\#\#\#\#\#\#\#\#\#\#\# \newline
        Reasoning: ... \newline
        Action: 0 / 1 / 2, ... \newline
        \#\#\#\#\#\#\#\#\#\#\#\#\#\#\#\#\#\#\#\#\#\#\#\#\#\#\#\#\#\#\#\#\#\#\# \\
        Response & Reasoning: Summary 2 provides more information about Michael giving a presentation to business students, making it more relevant to the question about who invited Michael to his business school as a guest speaker. \newline Action: 2 \\
        \midrule
        Node 2 & Read the text in triple quotes and answer a question: \newline
        Story background information: Michael and Ryan are on their way to give a speech at a business school. Kevin joins them after experiencing a tire blowout on his way there. Michael is excited about [...] \newline
        Main text: \newline 
        [SCENE\_BREAK] \newline
        Michael: So, you wanna start a business. How do you start? What do you need? Well, first of all, you need a building. And secondly, you need supply. You need something to sell. Now this could be anything. It could be... a... thingamajig. Or a... a whosi-whatsi. Or...  a Whatchamacallit . Now, you need to sell those in order to have a PayDay . And, if you sell enough of them, you will make a 100 Grand .  Satisfied?\newline
        [SCENE\_BREAK] \newline
        Toby: Oh, this looks great. I'd, I'd love to be there, but my daughter's play is tonight. ... Damnit! You know, one of the other parents will probably videotape it. \newline
        [...] \newline
        Question: "Who invited Michael to his business school as a guest speaker?"
        If the answer CANNOT be inferred from the text above, reply with action -1. \newline
        If the answer CAN be inferred from the text above, reply with action -2, and also provide your reasoning, and the final answer. \newline
        You are ONLY allowed to reply with action -2 or -1. \newline
        Your should reply with the following format: \newline
        \#\#\#\#\#\#\#\#\#\#\#\#\#\#\#\#\#\#\#\#\#\#\#\#\#\#\#\#\#\#\#\#\#\#\#
        \newline
        Reasoning: ... \newline
        Action: -2 or -1 \newline
        Answer: (A) ... \newline
        \#\#\#\#\#\#\#\#\#\#\#\#\#\#\#\#\#\#\#\#\#\#\#\#\#\#\#\#\#\#\#\#\#\#\# \\
        Response & Reasoning: The text does not explicitly mention who invited Michael to the business school as a guest speaker. \newline Action: -1 \\
        \midrule
        Node 1 & [...] // Retracted to the first node. \\
        Response & Reasoning: Summary 0 contains information about Michael, Ryan, and Kevin going to give a speech at a business school. This summary is the most relevant to the question about who invited Michael to the business school as a guest speaker. \newline
        Action: 0 \\
        \midrule
        Node 3 & [...] \newline 
        Main text: [...] \newline 
        Michael: Pop quiz. \newline 
        Kevin: ...What? \newline 
        Michael: Why is today a special day? \newline 
        Kevin: I almost died. \newline 
        Michael: Today's a special day, because I am being honored as a... visiting... professor, special lecturer, emeritus... how did you, how did you... \newline 
        Ryan: You will be a guest speaker... in my Emerging Enterprises class. \newline 
        Michael: In business school, Kevin. Business school. \newline 
        [...] \\
        \midrule
        Response & Reasoning: Ryan mentioned that Michael would be a guest speaker in his Emerging Enterprises class. \newline
        Action: -2 \newline
        Answer: Ryan \\
        \bottomrule
        \end{tabular}
    }
    \caption{
        Example trajectory from the SummScreenFD dataset. 
    }
    \label{table:extra_examples}
\end{table*}

\end{document}